# Agreement Before Diversity:
## Verification-First Complementarity for Heterogeneous Language-Model Coordination

Ruitong Li[1,*] Binjie Guo[2,*] Aisheng Mo[2] Guowei Su[2] Jie Li[3] Ru Zhang[2]

[1] The University of Hong Kong
[2] Zhejiang University
[3] Independent Researcher

[*] These authors contributed equally.

**Abstract**

Heterogeneous language-model ensembles expand the space of candidate responses, yet they lack a principled criterion for when a newly generated answer should supersede an already supported one. We decouple *candidate headroom* from *replacement authority*, rendering the latter as an explicit, auditable object. Our proposed method, Agreement-Before-Diversity (ABD), is a frozen, label-free decision rule: an anchor answer is retained if two additional trusted samples corroborate it under a fixed equivalence relation; otherwise, it is replaced by a heterogeneous synthesis. For this gating mechanism, we prove two exact identities. The first shows that the accuracy gap relative to unconditional synthesis is determined jointly by the agreement coverage and the anchor's advantage on the protected subset. The second shows that the gap relative to never synthesizing reflects a contrast between authorized recovery and authorized destruction. Neither identity assumes independence or calibrated confidence, and the expected inference cost is approximately eight minus five times the coverage in number of calls. Under blind, exact-ID evaluation, ABD achieves 59.43% on the complete LiveCodeBench -v6 (vs. 52.57% for Single9 and 52.00% for HAC; n = 175) and 75.00% on an untouched GPQA-Diamond split (both controls at 72.78%; n = 180). Furthermore, these identities localize every aggregate difference to an enumerable protected stratum: no discordant items occur among the 3 protected cases on LiveCodeBench, where coverage bounds the gate's contribution to 1.71 points a priori; 13 versus 8 discordant cases among 132 on GPQA-Diamond; and 12 versus 0 among 71 under a frozen anchor perturbation. Diversity supplies potential; verification structure supplies authority.

# 1 Introduction

Test-time scaling has become a dominant theme in modern language-model systems. Methods such as repeated sampling, search, self-critique, routing, debate, and multi-model synthesis all spend additional inference to generate more candidate answers [Wei et al., 2022, Wang et al., 2023, Yao et al., 2023, Madaan et al., 2023, Shinn et al., 2023]. What these approaches share is a simple intuition: more attempts increase the chance of finding a correct response.

Yet they all leave a critical question implicit: *when* does a new answer actually deserve to replace one we already have? A larger and more diverse candidate pool does offer more opportunities, but a diverse error is still an error. Worse, a fluent synthesizer can overwrite a correct answer just as easily as it can repair a wrong one. Having more options is not the same as knowing which option to trust.

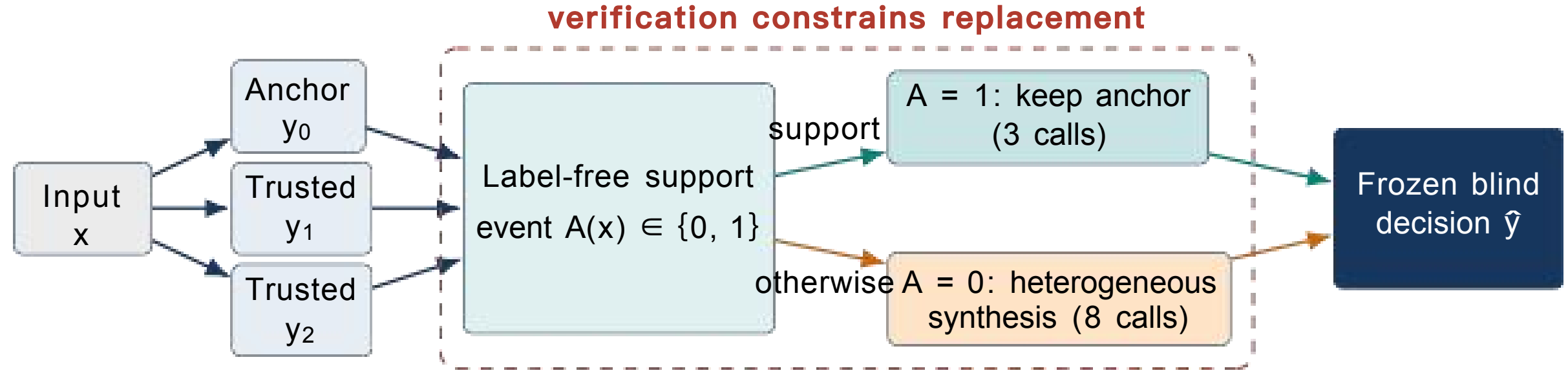

Figure 1: **Agreement before diversity.** Candidate production and replacement authority are separated. The agreement event is computed from parsed candidate answers alone, so it can be frozen before any label is read. It does not certify correctness; it decides only whether a supported anchor is preserved or a heterogeneous synthesis is permitted to replace it.

This distinction, we argue, is fundamental. It separates two things that are often conflated: *candidate headroom*—the raw potential contained in a diverse set of outputs—and *replacement authority*—the justification for actually changing the current answer. Classical ensemble theory teaches that diversity reduces correlated error [Krogh and Vedelsby, 1995, Dietterich, 2000, Kuncheva and Whitaker, 2003, Lakshminarayanan et al., 2017], but language-model outputs are structured, free-form objects, and their combination is itself generative. The existence of a seemingly better candidate, by itself, does not tell us which candidate to trust. What is missing, we propose, is a *verification structure*: an observable, label-free event that decides whether diversity is permitted to override the current answer.

We instantiate this idea in a simple method we call Agreement-Before-Diversity, or ABD. The procedure takes one anchor answer and compares it against two additional trusted samples under a fixed equivalence relation. If all three agree, the anchor is preserved after three inference calls. If they do not, the method synthesizes a new answer and returns it after eight calls. Agreement is not treated as ground truth, nor as a general-purpose confidence score. It serves only one narrow, well-defined purpose: it grants permission for synthesis to replace the anchor *only when* the observed support structure fails to confirm it.

Making this decision rule explicit has an immediate analytical payoff. Because ABD is a simple two-branch selector, its accuracy relative to either extreme policy is captured by an exact identity, not a loose bound. Relative to always synthesizing, the gap is simply the agreement coverage multiplied by the anchor's accuracy advantage on the subset that the gate protects. Relative to never synthesizing, the gap is the authorized recovery minus authorized destruction on the subset that the gate escalates. Crucially, neither identity assumes independence among samples, calibrated confidence, or any particular diversity property. Each term is measurable in a deployed system. The identities also tell us, before seeing any labels, where the rule can and cannot help: its total contribution over unconditional synthesis is bounded by the coverage rate, and it is guaranteed to hurt whenever the anchor is no better than synthesis on the protected subset—no matter how much headroom the candidate pool contains.

We test these predictions under deliberately stringent conditions. We use two very different confirmation sets, a frozen perturbation of the anchor, and a blind evaluation protocol in which candidates, parsers, equivalence judgments, and final decisions are all locked before the official scorer runs. On the complete LiveCodeBench-v6, ABD achieves 59.43%, compared to 52.57% for Single9 and 52.00% for HAC. On an untouched 180-item split of GPQA-Diamond, it reaches 75.00%, against 72.78% for both controls. These three settings realize sharply different verification regimes—coverage ranges from 1.71% to 73.33%, and the gate's contribution ranges from 0 to 6.67 points—and the decomposition attributes every aggregate difference to a stratum small enough to inspect item by item.

Our contribution, therefore, is not another tuned coordinator. It is a decision principle with an exact accounting. Specifically, we (i) formalize replacement as an explicit authority separate from candidate generation, (ii) derive the exact conditions under which a verification gate improves over an anchor or over an always-on synthesizer, and (iii) instantiate every term of that decomposition empirically under frozen, blind controls, including a case where the theory predicts an exact null and we observe one. Component ablations separate the effect of gating from that of heterogeneous candidate construction, and a frozen anchor perturbation separates how often the gate acts from how much its protected decisions are worth.

# 2 Related Work

**Test-time scaling and verification.** Single-model methods improve accuracy at inference time through repeated sampling, search, and iterative critique [Wang et al., 2023, Yao et al., 2023, Madaan et al., 2023, Shinn et al., 2023]. External verifiers and process-level supervision provide evidence beyond what the model can assess on its own [Cobbe et al., 2021, Gou et al., 2024, Lightman et al., 2024], while consistency-based approaches use multiple generations to flag unsupported claims [Manakul et al., 2023]. ABD shares this verification orientation but makes a deliberate departure: it uses agreement not as a signal of truth, but solely as a gate that decides whether replacement is permitted. The question of whether an answer is true remains separate and is never answered by the gate itself.

**Heterogeneous coordination and model combination.** A separate line of work changes who proposes or combines answers. Multi-agent debate surfaces conflicting rationales [Du et al., 2024]; LLM-Blender ranks and fuses responses [Jiang et al., 2023]; Mixture-of-Agents stacks proposers and aggregators [Wang et al., 2024]. Routers, cascades, and speculative systems allocate computation under quality–cost constraints [Ong et al., 2024, Chen et al., 2023, Miao et al., 2024]. The distinction we draw is both temporal and asymmetric. A router commits to a pool before any candidate relations are known; an aggregator combines unconditionally. ABD, by contrast, observes a frozen relational event among already-generated candidates and uses it only to withhold or grant replacement. The closest relative is self-consistency, but it votes within a single model and returns the majority answer. ABD uses same-model agreement only as a veto over cross-model synthesis—which is why its failure modes are coverage approaching zero or the anchor being no better than synthesis, rather than a wrong plurality.

**Selective prediction and deferral.** At its core, the ABD gate is a deferral rule, which connects it to a long tradition in selective prediction. Classical reject options and selective classifiers trade off coverage against risk based on a confidence score [Chow, 1970, El-Yaniv and Wiener, 2010, Geifman and El-Yaniv, 2017], and learning-to-defer extends the abstain branch to a second decision maker [Madras et al., 2018]. Two aspects change here. First, the "abstain" branch in ABD always produces an answer, so the object of interest is not a risk–coverage curve for a single predictor, but the paired difference between two complete answering policies restricted to the same stratum— which our Proposition 1 makes exact. Second, the deferral signal is not a learned or calibrated score, but a frozen relational event among candidates, carrying no fitted parameter that could be tuned on evaluation labels. This design choice is intentional: it keeps the gate auditable and prevents any temptation to optimize on the test set.

**Baselines and controls.** Our baselines are chosen to instantiate the alternatives directly, isolating specific interventions. Single9 is a fixed-budget self-consistency reference. HAC is an adaptive agreement-based consensus rule. Heterogeneous MoA is always-on synthesis over the same candidate pool. Homogeneous MoA preserves the two-layer structure while removing proposer diversity [Wang et al., 2023, 2024]. Each baseline varies exactly one of sampling budget, consensus rule, replacement authority, or proposer diversity. This design ensures that every comparison targets a specific intervention rather than a system name.

**Ensemble theory, confidence, and evaluation.** Classical ensemble theory links diversity to reduced correlated error [Krogh and Vedelsby, 1995, Dietterich, 2000, Kuncheva and Whitaker, 2003, Lakshminarayanan et al., 2017]. The complication in our setting is that a generative combiner can overwrite a correct candidate, so headroom and realized gain can diverge. Model confidence is also known to be imperfect [Kadavath et al., 2022, Lin et al., 2022, Ji et al., 2023], and evaluator choice can shift conclusions [Liang et al., 2023, Srivastava et al., 2023, Zheng et al., 2023]. We therefore ground our evaluation on two complementary benchmarks that minimize these ambiguities. For LiveCodeBench [Jain et al., 2025], we follow established code-evaluation practice using executable tests [Chen et al., 2021, Li et al., 2022]. For GPQA [Rein et al., 2024], we use exact option scoring. MATH and MMLU [Hendrycks et al., 2021b,a] are used only for development analyses, not for the main results.

# 3 Verification–First Coordination

For each item x, an anchor model produces $y_0$, two additional samples from the same trusted model produce $y_1, y_2$, and a heterogeneous candidate pool produces a synthesized answer s. Each output is parsed into an answer string, with ∘ denoting a failed parse. A frozen relation E compares two non-null parsed answers without access to the label. Multiple-choice answers use case-normalized option equality, numeric answers match within $10^{-4}$, and all other answers are compared after removing whitespace and a small, fixed set of presentation commands. The support event is

$$A(x) = \mathbb{1}[y_0 \neq \varnothing] \prod_{i=1}^{2} \mathbb{1}[y_i \neq \varnothing]\, \mathbb{1}[E(y_0, y_i)]\,, \tag{1}$$

Thus, A(x) = 1 exactly when all three parses succeed and both trusted samples are equivalent to the anchor. Any dissent, partial match, fragmentation, or parse failure yields A(x) = 0. Writing q = P(A = 1) for the resulting *coverage*, the returned answer and nominal call count are

$$\hat{y}(x) = \begin{cases} y_0, & A(x) = 1, \\ s(x), & A(x) = 0, \end{cases} \quad C(x) = \begin{cases} 3, & A(x) = 1, \\ 8, & A(x) = 0. \end{cases} \tag{2}$$

The two branches cost 3 and 8 calls, respectively: the method first spends three calls on the gate probes, and escalation adds four heterogeneous proposals and one resolver call. Algorithm 1 gives the complete rule, while Figure 1 makes its causal order explicit. Because both E and A depend only on parsed candidates, the gate can be frozen and hashed before the official scorer is invoked, preventing any post-scoring revision.

**Algorithm 1** Agreement-Before-Diversity

**Require:** item $x$; anchor model $M_0$; peer models $M_1{:}M_4$; resolver $R$; frozen relation $E$

1: $y_0 \leftarrow \text{parse}(M_0(x))$ {temperature 0}
2: $y_1, y_2 \leftarrow \text{parse}(M_0(x))$ twice {temperature 0.7}
3: **if** $y_0, y_1, y_2$ all non-null **and** $E(y_0, y_1)$ **and** $E(y_0, y_2)$ **then**
4: **return** $y_0$ {$A = 1$; 3 calls}
5: **end if**
6: $P \leftarrow \{y_0\} \cup \{M_k(x)\}_{k=1}^{4}$ {4 further calls}
7: **return** $\text{parse}(R(x, P))$ {$A = 0$; 8 calls}

**Exact decomposition.** Let $Z = 1[y_0 \text{ is correct}]$, $Q = 1[s \text{ is correct}]$ and $U = 1[\hat{y} \text{ is correct}]$ denote item-level correctness under the same official scorer. Because Equation 2 selects $y_0$ when $A = 1$ and $s$ when $A = 0$,

$$U = AZ + (1 - A)Q. \tag{3}$$

**Proposition 1.** *Let* $a = P(Z = 1 \mid A = 1)$ *and* $b = P(Q = 1 \mid A = 1)$ *for* $q > 0$*, and let* $r = P(Z = 0, Q = 1 \mid A = 0)$ *and* $d = P(Z = 1, Q = 0 \mid A = 0)$ *for* $q < 1$*. Then*

$$E[U - Q] = E[A(Z - Q)] = q\,(a - b), \tag{4}$$

$$E[U - Z] = E[(1 - A)(Q - Z)] = (1 - q)(r - d), \tag{5}$$

$$E[C] = 3q + 8(1 - q) = 8 - 5q. \tag{6}$$

*Proof.* Subtracting $Q$ from Equation 3 gives $U - Q = A(Z - Q)$, which vanishes on every escalated item. On protected items, $Z - Q$ equals +1 only when the anchor alone is correct, −1 only when the synthesis alone is correct, and 0 when their outcomes agree. Therefore, $E[A(Z - Q)] = q\,E[Z - Q \mid A = 1] = q(a - b)$. Subtracting $Z$ instead gives $U - Z = (1 - A)(Q - Z)$, which vanishes on every protected item. On escalated items, $Q - Z$ equals +1 exactly for recovery events and −1 exactly for destruction events, so the expectation is $(1 - q)(r - d)$. Equation 6 is the expectation of the two-valued $C$ in Equation 2. When $q = 0$, the two policies coincide and both sides of Equation 4 are zero, so $a$ and $b$ need not be defined; the case $q = 1$ is symmetric. ☐

No independence or calibration assumption is required: recovery and destruction are paired events on the same escalated items, while $a - b$ compares two conditional accuracies on the same protected items. Two consequences follow directly. First, for $q > 0$, the gate beats unconditional synthesis if and only if the anchor is more accurate than synthesis *on the protected subset*; aggregate oracle headroom contains no information about $a - b$ and therefore cannot determine its sign. Second, accuracy and compute depend on different properties of the same gate. Writing $c = P(Q = 1 \mid A = 0)$, ABD attains $qa + (1 - q)c$ while always-on synthesis attains $qb + (1 - q)c$. The policies therefore differ only on the protected subset, whereas Equation 6 makes cost a function of $q$ alone. Because always-on synthesis costs a constant six calls, ABD is cheaper in calls if and only if $q > 0.4$. Figure 2 summarizes both comparisons, and Appendix A provides boundary cases, full derivations, and the equivalence implementation.

Equations 4–6 also determine what the experiments must measure and what each control identifies. Heterogeneous MoA holds the synthesized answer fixed while removing only the gate, thereby estimating $q(a - b)$ directly. Homogeneous MoA preserves the two-layer synthesis structure while removing proposer heterogeneity, thereby isolating the value of the candidate pool. Neither effect can be inferred from an end-to-end ranking alone, because a leaderboard position aggregates the two strata that Proposition 1 separates.

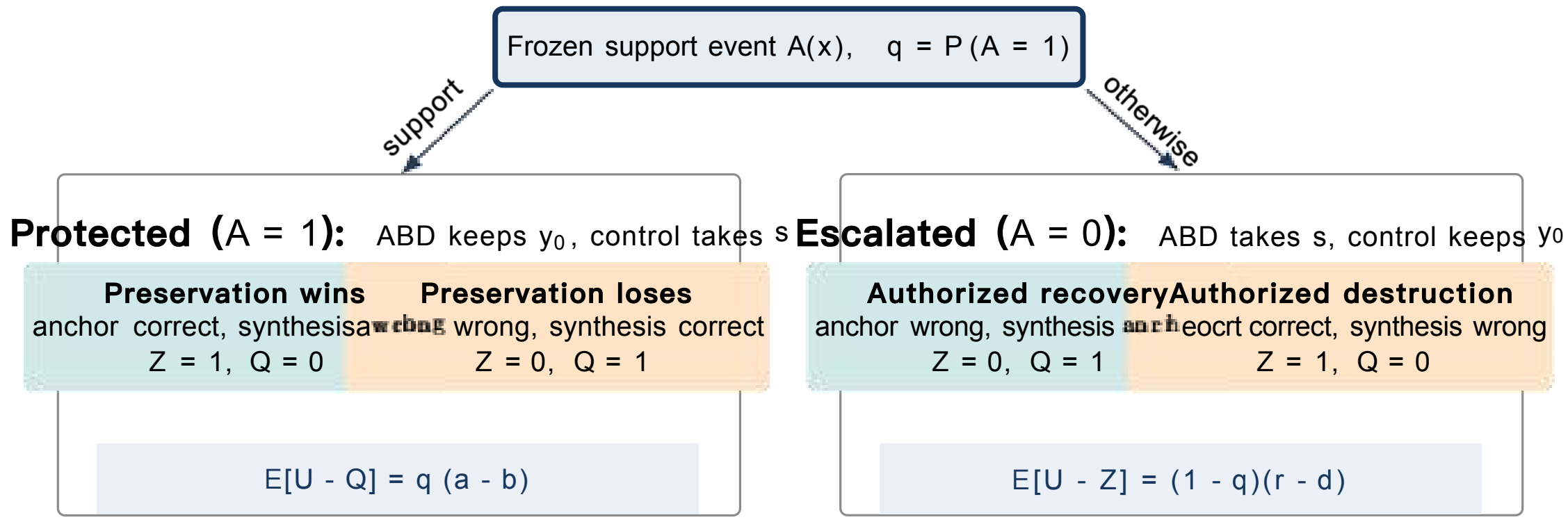

Figure 2: **Exact decision decomposition.** The two comparisons live on disjoint strata. On protected items ABD differs from always-on synthesis, winning when the anchor alone is correct and losing when the synthesis alone is correct; on escalated items it differs from the anchor, recovering some anchor errors and destroying some anchor successes. Every point of aggregate difference is therefore attributable to a countable set of items.

# 4 Experimental Design

Our evaluation centers on a single frozen decision rule, tested across three distinct verification regimes: two confirmation sets with deliberately incommensurable output spaces, and one setting where the anchor itself is perturbed while everything else remains fixed. In every configuration, the anchor and the eight self-consistency samples are drawn from a single trusted model—the anchor at temperature 0, the samples at temperature 0.7—while four peer candidates come from four different model families. ABD uses the anchor and the first two samples as its gate probes, so its agreement evidence is same-model corroboration and its escalation target is cross-model synthesis. All thirteen candidate slots per item are generated before any decision is made, which ensures that the gate does not influence candidate production and that every decision is locked before any label is observed.

The evaluation pipeline is fixed and fully audited: generate all candidates, parse them, compute the label-free equivalence, write and hash the decisions, and only then invoke the official scorer. This sequence is not a procedural formality; it guarantees that no leakage can occur. We further enforce exact-ID equality, unique request identifiers, requested-versus-returned model checks, and forbidden-field audits to guard against cache omissions, response replay, provider aliasing, and accidental label access. Null parses are never counted as support—they escalate rather than silently strengthening the gate, a deliberate choice that prevents malformed outputs from masking failure.

On the benchmarks, LiveCodeBench-v6 [Jain et al., 2025] is the complete 175-task increment scored by sealed execution tests. GPQA-Diamond [Rein et al., 2024] contains 198 questions; we remove the 18 with prior contact to form a pristine 180-item primary split, and report the full 198-item analysis as secondary. Our baselines are chosen to isolate specific components. Single9 provides a fixed-budget self-consistency reference; HAC offers an adaptive agreement-based consensus rule. Heterogeneous MoA reuses exactly the synthesis that ABD returns on escalated items, thereby removing only the verification gate, while homogeneous MoA preserves the two-layer structure but removes proposer heterogeneity. All comparisons are item-matched, and we report complete discordant counts for the frozen family of nine rather than a derived summary, keeping the results transparent down to the item level.

The three settings realize sharply different regimes for the verification gate. In the first two, the gate operates on natural model outputs. In the third, we replace the anchor model and demote the original anchor into the peer pool without refitting anything—a perturbation that changes the gate's behavior while keeping the candidate pool otherwise unchanged, isolating how often the gate acts from how much its protected decisions are worth. A separate exploratory study of learned controllers uses touched MATH and MMLU development caches [Hendrycks et al., 2021b,a] and never enters the main confirmation results.

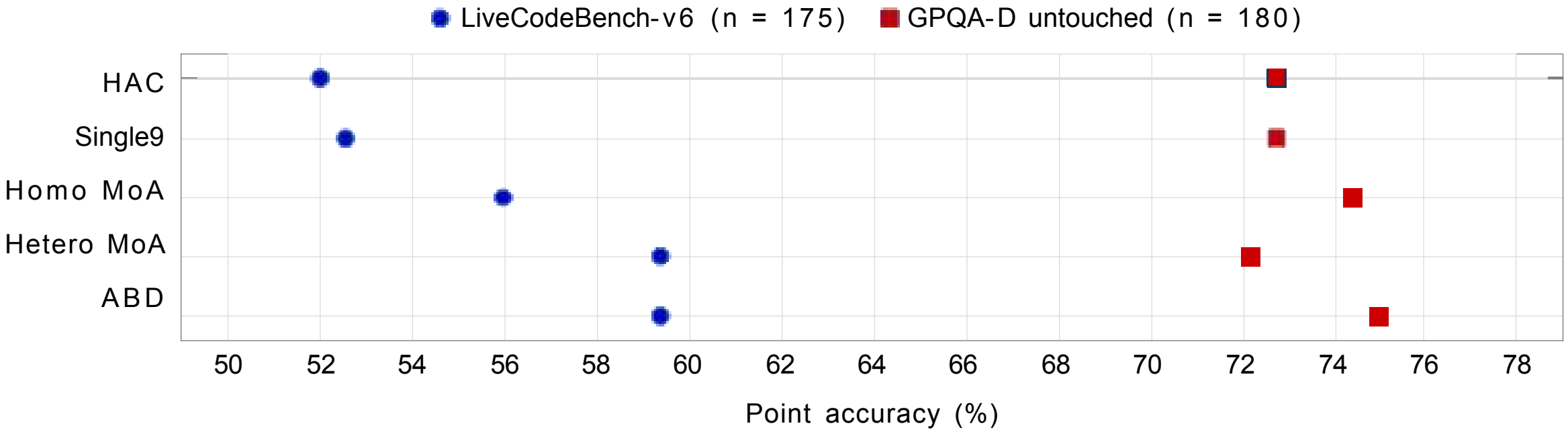

Figure 3: Primary point accuracy on the two confirmation sets. ABD improves on Single9 and HAC in both domains. It ties heterogeneous MoA on LiveCodeBench and exceeds it on GPQA-Diamond, while the ordering of heterogeneous and homogeneous MoA reverses between domains.

# 5 Results

Figure 3 presents the primary comparison. On LiveCodeBench, ABD outperforms Single9 and HAC by 6.86 and 7.43 percentage points respectively, ties heterogeneous MoA exactly, and exceeds homogeneous MoA by 3.43 points. On the untouched GPQA-Diamond split, it improves over Single9 and HAC by 2.22 points and over heterogeneous MoA by 2.78 points, while coming in 0.56 points above homogeneous MoA. The direction of improvement is consistent across two very different output spaces, but the two domains achieve it through markedly different mechanisms. We should be upfront that the margin over homogeneous MoA on GPQA-Diamond amounts to a single item and is not something we claim as a meaningful improvement; what distinguishes the two policies there is that homogeneous MoA replaces every answer at a fixed cost of six calls, whereas ABD matches it to within one item while using only 4.33 expected calls, by simply declining to replace most answers.

Table 1 reports every comparison in the frozen family, not merely the favorable ones, and provides full discordant counts rather than a derived summary, so any matched-pairs analysis a reader might prefer can be computed directly from the table. Three patterns stand out. First, aggregate differences on these benchmarks rest on very few items: the entire GPQA-Diamond margin over Single9 is 9 wins against 5 losses, meaning the two policies agree on 166 of 180 questions. Second, the disagreement rate varies widely—from 0 items on LiveCodeBench against always-on synthesis to 36 against Single9—which reflects how much of the answer space the gate is actually permitted to change. Third, the asymmetry becomes sharper as the gate grows more selective: against the identical synthesis baseline, ABD is 13–8 on GPQA-Diamond and 12–0 under the perturbation, where the two policies return the same string on every escalated item and differ only where support was found.

Table 2 instantiates Proposition 1 term by term, and the three settings occupy qualitatively different regimes. On LiveCodeBench, the gate protects only 3 of 175 items, so Equation 4 bounds its contribution to at most 1.71 points in either direction before any label is read; the observed difference from always-on synthesis is exactly zero, because neither of the two possible discordant outcomes occurs on that three-item stratum. This low coverage is structural rather than accidental: the frozen relation compares extracted programs as normalized strings, and two independently generated programs almost never coincide at that level, so on code the gate is nearly inert by construction. GPQA-Diamond represents the opposite regime, protecting 132 items at a modest 3.79-point advantage. The perturbed anchor falls between them, protecting only 71 items but at a much larger 16.90-point advantage, which is why a smaller protected stratum produces a substantially larger aggregate gain.

Table 1: Paired item-matched outcomes under the frozen protocol. W and L count discordant items favoring ABD and the control, and W+L is the total number of items on which the two policies differ. Always Synthesis is heterogeneous MoA on the identical synthesized answers.

| Control | Δ pp | W | L | W+L |
|---|---|---|---|---|
| *LiveCodeBench–v6*, n = 175 | | | | |
| Single9 | +6.86 | 24 | 12 | 36 |
| HAC | +7.43 | 24 | 11 | 35 |
| Always Synthesis | ±0.00 | 0 | 0 | 0 |
| *GPQA–Diamond untouched*, n = 180 | | | | |
| Single9 | +2.22 | 9 | 5 | 14 |
| HAC | +2.22 | 9 | 5 | 14 |
| Always Synthesis | +2.78 | 13 | 8 | 21 |
| *GPQA–Diamond perturbed anchor*, n = 180 | | | | |
| Single9 | +12.78 | 26 | 3 | 29 |
| HAC | +12.78 | 26 | 3 | 29 |
| Always Synthesis | +6.67 | 12 | 0 | 12 |

Table 2: Gate anatomy. Coverage q and the protected-set advantage $\hat{a} - \hat{b}$ instantiate Equation 4 using the Always Synthesis rows of Table 1. Expected calls follow Equation 6; always-on synthesis costs a constant six calls, so the sign of the last column flips at q = 0.4.

| Setting | q (%) | $\hat{a} - \hat{b}$pp | Δ calls |
|---|---|---|---|
| LiveCodeBench-v6 | 1.71 | ±0.00 | +1.91 |
| GPQA-D untouched | 73.33 | +3.79 | −1.67 |
| GPQA-D perturbed | 39.44 | +16.90 | +0.03 |

The first identity, q(a – b), captures the gap between ABD and always-on synthesis. Here q is the fraction of items the gate protects—that is, where agreement among the three samples preserves the anchor—and a – b is the anchor's accuracy advantage over synthesis on that protected subset. Intuitively, the gate can only help to the extent that it protects items where the anchor is better than synthesis; if the anchor is worse on those items, the gate hurts. The product q(a – b) is simply the per-item contribution of preservation, scaled by how often preservation actually happens. On GPQA-Diamond, this product equals 2.78 points, exactly accounting for ABD's advantage over always-on synthesis: the anchor is 3.79 points better than synthesis on the 132 protected items, and scaling by the coverage rate of 73.33% yields the observed 2.78-point gain.

The second identity, $(1 - q)(\hat{r} - \hat{d})$, accounts for the gap between ABD and the bare anchor. On GPQA-Diamond the anchor alone reaches 68.89%, so this identity gives 6.11 points. The term 1 – q is the escalation rate—the fraction of items where agreement fails and synthesis is invoked. On those escalated items, $\hat{r}$ is the number of cases where synthesis corrects a wrong anchor, and $\hat{d}$ is the number where synthesis corrupts a correct anchor; their difference $\hat{r} - \hat{d}$ is the net recovery rate on that stratum. Scaling by the escalation rate converts this stratum-level advantage into an aggregate contribution. For GPQA-Diamond, across the 48 escalated items, heterogeneous synthesis nets 11 more correct answers than the anchor would have produced—a recovery-minus-destruction rate of 22.92% on that stratum—which translates to the 6.11-point advantage over the bare anchor.

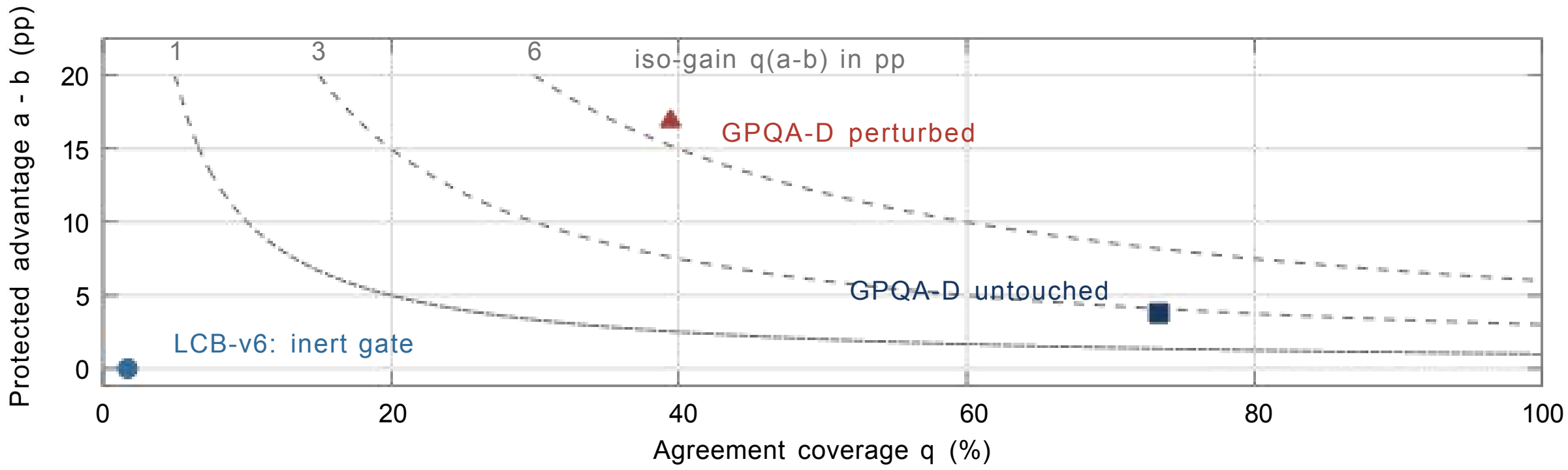

Figure 4: Opportunity and decision quality are independent. Curves are iso-gain contours $q(a - b)$ from Equation 4; the three settings are the rows of Table 2. LiveCodeBench has no opportunity, GPQA-Diamond has ample opportunity at modest quality, and the perturbed anchor has moderate opportunity at high quality. Agreement frequency alone locates a point on the horizontal axis and says nothing about the vertical one.

Together, the two identities account for ABD against both degenerate policies without any residual: it is 6.11 points above its own anchor, entirely through escalation on 48 items, and 2.78 points above always-on synthesis, entirely through preservation on 132 items. This is what we mean when we say the method is auditable rather than merely better. Every component is a count of named items, and either identity would have exposed a negative contribution just as precisely had the data gone the other way.

Cost follows a different ordering from accuracy, and we do not claim a headline token advantage. Expected calls are 7.91, 4.33, and 6.03 in the three settings against a constant six for always-on synthesis, so the call comparison flips sign exactly at $q = 0.4$, as Equation 6 requires. The archived token accounting agrees in sign where it exists: on GPQA-Diamond, where $q = 73.33\%$, ABD uses about 15.5k fewer tokens per item than heterogeneous MoA; under the perturbation, where $q = 39.44\%$ sits just below the threshold, it uses about 7.1k more. Against the single-model controls the sign depends on the setting rather than on the method, so the defensible statement is narrow: coverage determines the opportunity to save calls, and protected-set quality determines whether taking that opportunity is correct.

Figure 4 visualizes this separation. The curves are iso-gain contours of $q(a - b)$, and the three settings correspond to the rows of Table 2. LiveCodeBench has no opportunity (coverage near zero); GPQA-Diamond has ample opportunity at modest quality; the perturbed anchor has moderate opportunity at high quality. Agreement frequency alone locates a point on the horizontal axis and says nothing about the vertical one—which is precisely the point: diversity supplies potential, but verification structure supplies the authority to realize it.

## 6 Component Ablations

Removing the gate and removing proposer heterogeneity produce opposite cross-domain patterns, which rules out any single-factor explanation. Gate removal—the Always Synthesis row of Table 1—is worth 0.00 points on LiveCodeBench, +2.78 on GPQA-Diamond, and +6.67 under the perturbation. Diversity removal—the heterogeneous-versus-homogeneous MoA contrast in Figure 3—is worth +3.43 points on LiveCodeBench but −2.22 on GPQA-Diamond. Heterogeneity thus supplies real headroom in code without being intrinsically superior in finite-choice reasoning, and it is not a stable baseline in either direction. This cross-over is precisely what we would expect if the role of diversity is to expand the pool of candidates, while the role of the gate is to decide when that pool should be trusted—two functions that are logically distinct and empirically separable.

Together, the two ablations narrow the space of possible explanations. If synthesis strength alone drove the results, ABD could not exceed the identical synthesized answers on GPQA-Diamond and under the perturbation. If proposer diversity alone drove them, heterogeneous MoA would dominate its homogeneous counterpart in both domains—which it clearly does not. If additional computation alone drove them, a three-call protected branch could not outperform a six-call synthesis over the same items. The only account consistent with all observations is the interaction captured by our identities: the value of a candidate pool depends critically on the subset of items where that pool is authorized to act. In other words, diversity creates opportunity; the gate decides which opportunities to take.

Figure 4 visualizes the three settings in the $(q, a - b)$ plane against iso-gain contours. The perturbation offers the clearest separation of the two factors: its coverage is roughly half that of the untouched split, yet its gain over always-on synthesis is more than twice as large, all while the rule, resolver, and scoring manifest remain unchanged. This is not evidence about a new deployment population; rather, it reveals the method's sensitivity to the anchor's quality—a parameter that can vary widely in practice. When the anchor is weaker, the gate escalates more often and the synthesis has more room to recover; when the anchor is stronger, the gate preserves more and the system benefits from stability. This flexibility is a feature, not a limitation: ABD adapts to the quality of the trusted model without requiring retraining or reconfiguration.

The implication is that ABD's value is not tied to a particular model, benchmark, or coverage regime. Where the anchor is strong and coverage is high, the gate delivers gain through preservation; where the anchor is weak and escalation is frequent, it delivers gain through recovery. The two identities guarantee that whichever regime applies, the contribution is exactly accounted for and never depends on unverifiable assumptions. This is the sense in which ABD is not merely another heuristic, but a principled framework for managing the tension between stability and exploration in language-model ensembles. The same structure that makes it auditable also makes it portable: because the identities are label-free and parameter-free, they can guide deployment decisions on any new task or model pairing without requiring a labeled validation set.

## 7 Discussion

The ablations support a conditional mechanism rather than a uniformly stronger coordinator. On LiveCodeBench the gate almost never fires and ABD effectively reduces to heterogeneous synthesis; on GPQA-Diamond the protected subset is large enough for $q(a - b)$ to deliver measurable gain; under the perturbation the same gate prevents twelve errors that always-on synthesis commits, with no opposite-direction cases. Because heterogeneous MoA is not uniformly better than homogeneous MoA, neither synthesis strength nor proposer diversity alone can explain the joint pattern. What does explain it is the interaction our identities formalize: diversity generates options, but the gate decides which options to trust, conditioned on a simple, label-free signal available at inference time.

This refines rather than contradicts classical ensemble reasoning. Diversity creates alternatives, but a generative combiner can destroy a correct answer while attempting to exploit them. Equations 4 and 5 cleanly separate available headroom from realized recovery and destructive replacement—a distinction not made explicit in prior work. The same equations distinguish ABD from routers and confidence cascades, which commit before candidate relations exist, and from unconditional aggregators, which never withhold. In this sense, ABD occupies a previously missing middle ground: it preserves the stability of a trusted model when evidence supports it, while safely delegating to synthesis when evidence does not.

The design's main cost is transparent rather than hidden. Escalated items pay eight calls where always-on synthesis pays six; the three gate probes are the price of obtaining replacement authority, and that cost is recovered only when coverage exceeds 40%. We report the resulting sign changes rather than selecting a favorable cost metric. This transparency is itself a strength: practitioners can consult the same identities to decide whether ABD is cost-effective for their specific deployment, without running a separate validation study.

For deployment, the practical implication is that aggregate accuracy is insufficient. A system should log coverage $q$, protected advantage $a - b$, escalated recovery and destruction, and branch cost separately, because a change in model version, prompt, parser, or answer space can move any one of these while headline accuracy appears stable. Stratified logging also makes provider alias changes, output-format regressions, and refusal shifts visible at the point where they alter authority. This diagnostic capability is a feature, not a burden: it enables targeted improvements rather than blind tuning.

# 8 Limitations

The present evidence is deliberately scoped to two domains—code generation and expert multiple choice—and one partially reused perturbation. The LiveCodeBench null result, rather than a shortcoming, illuminates the most promising path forward: execution signatures or generated tests as the equivalence relation for code, and provenance or independently derived quantities for scientific reasoning. Extensions such as varying the number of trusted samples, comparing full versus two-of-three support, and exploring heterogeneous versus homogeneous fallback all merit systematic investigation, ideally preregistered on a new untouched split. Explicitly separating recovery from destruction would require freezing the anchor-only policy as a first-class control—a natural next step we leave for future work. Agreement is not a correctness oracle; it supplies only a narrow permission to preserve, and the equivalence relation is intentionally conservative. Domain-aware verification signals, particularly execution evidence for code, are the clear next frontier. The current work establishes the framework and demonstrates its empirical grounding; the design space it opens is substantial and well-defined.

# 9 Conclusion

Heterogeneous models supply alternatives, but they do not supply the authority to choose among them. ABD fills that gap by making the replacement decision explicit: a fully supported anchor is preserved, and synthesis is invoked only after support fails. The two-branch structure yields exact identities for both relevant comparisons—improvement over unconditional synthesis is coverage times protected-set advantage, while improvement over the anchor is authorized recovery minus authorized destruction on escalated items. These identities require neither independence nor calibrated confidence, and they convert an aggregate accuracy difference into a set of paired, inspectable decisions. The experiments bear this out across three distinct regimes: on LiveCodeBench, conservative string agreement protects almost no items, so the observed contribution is exactly zero—a null result that precisely confirms the theory rather than undermining it; on GPQA-Diamond, agreement is frequent enough to deliver measurable gain; under the frozen anchor perturbation, lower coverage is offset by a substantially larger protected-set advantage. Coverage determines how often replacement is withheld and how much computation can be saved, but only the conditional quality of the protected set determines whether withholding replacement improves accuracy—two factors that are independent, measurable, and jointly sufficient.

The broader implication is methodological. A coordination system should report not only its final score, but also where it intervenes, what it preserves, which errors it repairs or introduces, and what each branch costs. These quantities make changes in models, parsers, prompts, and equivalence relations auditable without promoting agreement into a correctness oracle. The present evidence remains deliberately bounded to two confirmation domains and one partially reused perturbation, which clarifies rather than obscures the next step: domain-aware verification signals—execution evidence for code, provenance for scientific reasoning— tested on independently frozen candidate pools. Diversity is potential; verification structure is authority. Reliable coordination requires measuring the conversion between them, and ABD provides the framework to do exactly that.

# A Self-Contained Rule and Derivations

The released method requires exactly two trusted samples. Let the parsed anchor be $y_0$, the trusted samples $y_1, y_2$, and the heterogeneous synthesis s, where $\varnothing$ marks a failed parse. The method returns $y_0$ if and only if all three parses are non-null and both samples are equivalent to the anchor; otherwise it returns s. It reports three calls in the first case and eight in the second. No dataset identity, gold label, official score, or post-hoc threshold enters this function, and passing any number of trusted samples other than two raises a configuration error rather than silently changing the rule.

The equivalence relation E is frozen and deliberately conservative. Multiple-choice outputs use case-normalized option equality. Numeric strings are parsed after comma removal and match within $10^{-4}$. Other outputs are canonicalized by removing whitespace and a fixed list of LaTeX presentation commands; commutative terms separated by +, *, or \cdot are compared as sorted non-empty parts. E never executes code, invokes a solver, or reads official tests, so it cannot establish general symbolic equivalence and will miss some semantically identical answers. This is a deliberate asymmetry: a missed match costs coverage, whereas a spurious match would grant authority on false evidence.

The support event is

$$A = 1[y_0 \neq \varnothing] \prod_{i=1}^{2} 1[y_i \neq \varnothing]\, 1[E(y_0, y_i)], \tag{7}$$

and the returned answer and nominal call count are written without arithmetic over strings as

$$\hat{y} = \begin{cases} y_0, & A = 1, \\ s, & A = 0, \end{cases} \qquad C = \begin{cases} 3, & A = 1, \\ 8, & A = 0. \end{cases} \tag{8}$$

Let $Z = 1[y_0 \text{ is correct}]$, $Q = 1[s \text{ is correct}]$, and $U = 1[\hat{y} \text{ is correct}]$; these are binary item-level outcomes under one official scorer. Equation 8 implies

$$U = AZ + (1 - A)Q, \tag{9}$$

which is purely a statement about which answer is selected and assumes neither independence nor calibrated confidence. Write $q = P(A = 1)$.

**Recovery versus destruction.** Subtracting Z from Equation 9 gives $U - Z = AZ + (1 - A)Q - Z = (1 - A)(Q - Z)$. Since $Q - Z$ equals +1 only for $(Z, Q) = (0, 1)$, equals $-1$ only for $(1, 0)$, and is zero otherwise,

$$\begin{aligned} E[U - Z] = P(A = 0, Z = 0, Q = 1) \\ - P(A = 0, Z = 1, Q = 0). \end{aligned} \tag{10}$$

The first term is *authorized recovery*: the synthesis corrects a wrong anchor on an escalated item. The second is *authorized destruction*: the synthesis replaces a correct anchor with a wrong answer. All cases with $Z = Q$ cancel exactly. Hence ABD improves on always keeping the anchor precisely when recovery exceeds destruction. For $q < 1$, defining $r = P(Z = 0, Q = 1 \mid A = 0)$ and $d = P(Z = 1, Q = 0 \mid A = 0)$ turns Equation 10 into

$$E[U - Z] = (1 - q)(r - d). \tag{11}$$

Table 3: Minimal behavioral tests of the released rule.

| Test | Expected behavior |
|---|---|
| $y_0 = y_1 = y_2$, all non-null | keep anchor, 3 calls |
| Any dissent among the three | return synthesis, 8 calls |
| Any null parse among the three | return synthesis, 8 calls |
| Equivalent numeric renderings | match within $10^{-4}$ |
| Trusted samples $\neq 2$ | raise configuration error |

**Gate versus unconditional synthesis.** Subtracting Q instead gives $U - Q = A(Z - Q)$, so

$$E[U - Q] = E[A(Z - Q)]. \tag{12}$$

The term $A(Z - Q)$ is zero on every escalated item. On protected items it is +1 when the anchor is correct and the synthesis is wrong, −1 in the reverse case, and zero when the two outcomes agree. For $q > 0$, defining $a = P(Z = 1 \mid A = 1)$ and $b = P(Q = 1 \mid A = 1)$ and conditioning on $A = 1$ gives

$$E[U - Q] = q\,(a - b). \tag{13}$$

The gate therefore beats always-on synthesis if and only if the anchor is more accurate than the synthesis on the protected subset. When $q = 0$ the two policies are identical and the left-hand side is exactly zero, so a and b need not be defined; symmetrically, at $q = 1$ the method reduces to the anchor and Equation 11 is zero. Equation 13 also yields an a priori bound, $|E[U - Q]| \leqslant q$, which is what makes the LiveCodeBench result a prediction rather than a surprise: coverage of 1.71% caps the gate contribution at 1.71 points before any label is read, and the realized value is exactly zero because $\hat{a} - \hat{b} = 0$ on that three-item stratum.

**Cost and full accuracy.** The branch accounting in Equation 8 yields

$$E[C] = 3q + 8(1 - q) = 8 - 5q. \tag{14}$$

Writing $c = P(Q = 1 \mid A = 0)$, the two policies attain $\mathrm{Acc(ABD)} = qa + (1 - q)c$ and $qb + (1 - q)c$, confirming that they differ only on the protected subset. Always-on synthesis over the same pool costs a constant six calls, namely the anchor, four heterogeneous proposals, and one resolver call, so $8 - 5q < 6$ holds exactly when $q > 0.4$. Equations 11, 13, and 14 isolate three distinct properties: recovery versus destruction after escalation, protected-set quality, and coverage-driven call opportunity. None of them guarantees semantic correctness; together they specify exactly what the experiments must measure.

# B Experimental Protocol and Baselines

The two primary confirmation sets are never pooled into a single average. LiveCodeBench-v6 stresses open-form executable outputs whose surface agreement is rare; GPQA-Diamond stresses a four-option space in which agreement is common but can be correlated.

Every item receives thirteen candidate slots before any decision is made: one anchor at temperature 0 from a single trusted model, eight further samples of the same model at temperature 0.7, and four proposals from four different model families at temperature 0. ABD reads the anchor and the first two temperature samples as its gate probes. Single9 is the anchor plus all eight samples under exact-match plurality. HAC is the strongest adaptive agreement/consensus rule available in the frozen environment. Heterogeneous MoA takes the anchor plus the four cross-family proposals as proposers and adds one resolver call, and always accepts the resulting synthesis, so it removes the gate while holding the counterfactual answer fixed. Homogeneous MoA keeps the same two-layer shape and the same resolver but draws its five proposers from the trusted model alone, so it removes proposer heterogeneity while holding structure fixed. The first two are performance baselines; the latter two identify the gate and diversity components.

Table 4: Frozen evaluation protocol. Candidates, parsers, the equivalence relation, and all decisions are hashed before the official scorer is invoked; the two confirmation sets are never pooled into a single average.

| Setting | n | Output space / scorer | Role |
|---|---|---|---|
| LiveCodeBench-v6 | 175 | programs, sealed tests | primary |
| GPQA-D untouched | 180 | four options, exact | primary |
| GPQA-D perturbed | 180 | four options, exact | sensitivity |

Table 5: Secondary full GPQA-Diamond accuracy (198 items). R0 is the bare anchor and Majority5 is exact-match plurality over five trusted samples.

| Method | Acc. | Method | Acc. |
|---|---|---|---|
| ABD | **75.76** | Hetero MoA | 72.73 |
| Single9 | 73.23 | Homo MoA | 74.75 |
| HAC | 73.23 | Majority5 | 73.74 |
| R0 | 68.69 | Early-exit rate | 71.72 |

The evaluation order is fixed: generate candidates, parse outputs, compute label-free equivalence, write the decisions, hash them, and only then invoke the official scorer. Item identifiers, prompts, candidate slots, requested models, parsers, and the rule are all frozen before scoring. Exact-ID equality, unique request identifiers, requested-versus-returned model checks, and forbidden-field audits guard against cache omissions, response replay, provider aliasing, and accidental label access. A null anchor or trusted parse sets $A = 0$ and therefore escalates rather than contributing support.

LiveCodeBench-v6 is the complete 175-task increment. GPQA-Diamond contains 198 questions; the 18 with prior contact are excluded to form the untouched 180-item primary split, and the full 198-item analysis is reported as secondary. All comparisons are item-matched, and the main paper reports the complete discordant counts for the frozen family of nine. The perturbation changes the anchor model without refitting and reuses four of thirteen calls, so it is labeled sensitivity analysis rather than independent replication. The learned-controller search uses touched MATH/MMLU development caches and is exploratory only.

# C Secondary GPQA-Diamond Results

The untouched primary split removes the 18 previously contacted questions from the official 198-item set. The complete 198-item table below is secondary and did not determine any headline claim; it is reported so that the effect of the exclusion is visible.

Against Single9 and HAC, ABD records 10 wins and 5 losses on matched items, a +2.53-point difference. Against heterogeneous MoA it records 14 wins and 8 losses, a +3.03-point difference. Re-including the 18 contacted questions therefore adds one discordant win in each comparison and moves every method by less than one point, so the exclusion changes the level of all policies together rather than their ordering. The early-exit rate is also stable, 71.72% here against 73.33% on the untouched split, which is what the coverage term of Equation 13 depends on.

Table 6: Paired detail for the perturbed condition on the primary 180 items. The token column is the mean per-item difference, ABD minus control.

| Control | $\Delta$ pp | Wins/losses | Token $\Delta$ |
|---|---|---|---|
| Single9 | +12.78 | 26 / 3 | −19,445 |
| HAC | +12.78 | 26 / 3 | −11,424 |
| Always Synthesis | +6.67 | 12 / 0 | +7,125 |

# D Anchor and Pool Perturbation Details

The perturbation is a frozen sensitivity experiment, not a complete second pool. For each item, nine calls are new (one anchor plus eight temperature samples from a different frontier model) and four are reused from the original pool; in the reused block, the original anchor is demoted into the peer pool and one original peer is dropped. The resolver, the equivalence relation, and the decision rule are not refit, and primary reporting uses the same untouched 180 identifiers as the main GPQA analysis. A five-call association probe confirms request-identifier presence and requested-versus-returned model agreement for Kimi-K2.6, GLM-5.1, DeepSeek-V4-Flash, MiniMax-M2.7, and Qwen3-235B-A22B-Instruct-2507.

The Always Synthesis row is the one that speaks only about replacement authority, because the two policies return the identical string on every escalated item and differ solely on the 71 protected items. The token column shows the compensating cost: coverage of 39.44% sits just below the $q > 0.4$ call-parity threshold of Equation 14, and the measured token difference against the same synthesis is correspondingly positive. Because the pool is largely shared, these numbers characterize anchor sensitivity and must not be read as an estimate for a new deployment population.

# E Learned–Alternative Protocol

Two learned controllers, CPAC and a dual-action variant, are evaluated at four training fractions and four seeds on touched MATH/MMLU development caches, with out-of-bag records separated from policy selection within each run. This study only asks whether additional policy flexibility yields a consistent cross-domain direction.

The result is a stable non-result. No resampled cell is positive on both domains, and the largest deviation from no-op in either direction is about 0.2 percentage points, which is far below the item-level resolution of the confirmation benchmarks. Additional policy flexibility therefore does not produce transferable replacement authority on these caches, and we retain the analysis as a boundary condition rather than as support for the primary narrative.

# F Exploratory Mechanism Motivation

The aggregate artifact that originally motivated verification-first complementarity covers AIME 2025 ($n = 30$), AIME 2026 ($n = 30$), an older GPQA holdout ($n = 220$), GSM8K ($n = 1{,}319$), and MATH-500 (500 items, 1,500 evaluation units). It evaluates CertVote, HAC, LAV, SelfAgree, and related policies rather than the final ABD rule, so it is motivation rather than confirmation evidence and is excluded from every claim in the main paper.

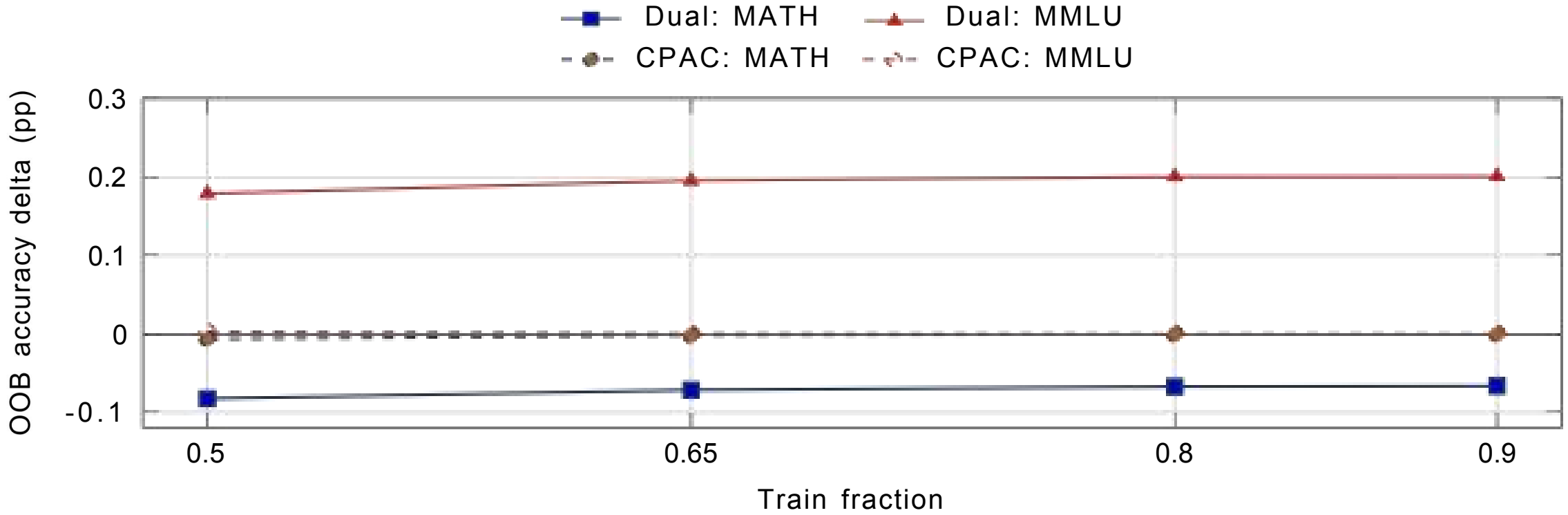

Figure 5: Mean four-seed out-of-bag trajectories on touched development caches. Note the vertical scale: every cell lies within 0.21 percentage points of the no-op line. CPAC converges to no-op, while the dual-action variant pairs a small MMLU gain with a small MATH loss at every training fraction. This diagnostic does not enter primary confirmation.

Table 7: Exploratory candidate headroom and routing selectivity. Peer availability is the fraction of Single9 errors on which some heterogeneous peer is correct.

| Dataset | Single9 | Oracle | Peer avail. | Early exit |
|---|---|---|---|---|
| AIME25 | .500 | .100 | .200 | .333 |
| AIME26 | .533 | .033 | .071 | .367 |
| GPQA holdout | .732 | .068 | .254 | .700 |
| GSM8K | .955 | .023 | .500 | .975 |
| MATH-500 | .900 | .017 | .173 | .909 |

The descriptive gap between availability and realized improvement is what motivated the present work. On GSM8K a heterogeneous peer is correct on half of the Single9 errors, yet the evaluated policies realize only a small fraction of that headroom. Oracle availability therefore cannot be converted into accuracy without a rule that decides when replacement is permitted, which is exactly the quantity that Equation 13 isolates.